\RequirePackage{fix-cm}
\documentclass[twocolumn]{svjour3}          
\smartqed  
\usepackage{graphicx}
\usepackage{amsmath,amssymb} 
\usepackage{color}
\usepackage{dsfont}
\usepackage[algo2e]{algorithm2e}
\usepackage{algorithmic}
\usepackage{algorithm}
\usepackage{subfigure}
\usepackage{booktabs}
\usepackage{multirow}
\usepackage[pagebackref=true,breaklinks=true,letterpaper=true,colorlinks,bookmarks=false,citecolor=blue]{hyperref}
\usepackage{cite}
\usepackage{natbib}
\usepackage{bbm}

\begin{document}

\title{Twin Contrastive Learning for Online Clustering}


\author{Yunfan Li \and
        Mouxing Yang \and
        Dezhong Peng \and
        Taihao Li \and
        Jiantao Huang \and
        Xi Peng
\thanks{Corresponding author: Xi Peng.}
}


\institute{Y. Li, M. Yang, D. Peng, and X. Peng \at
               College of Computer Science, Sichuan University. 
              Chengdu, China.\\
              \email{\{yunfanli.gm, yangmouxing, pengx.gm\}@gmail.com, pengdz@scu.edu.cn}
              \and
              T. Li and J. Huang \at
              Zhejiang Lab. Hangzhou, China.\\
              \email{lith@zhejianglab.com, jthuang@zhejianglab.edu.cn}
}

\date{Received: date / Accepted: date}

\maketitle

\begin{abstract}
This paper proposes to perform online clustering by conducting twin contrastive learning (TCL) at the instance and cluster level. Specifically, we find that when the data is projected into a feature space with a dimensionality of the target cluster number, the rows and columns of its feature matrix correspond to the instance and cluster representation, respectively. Based on the observation, for a given dataset, the proposed TCL first constructs positive and negative pairs through data augmentations. Thereafter, in the row and column space of the feature matrix, instance- and cluster-level contrastive learning are respectively conducted by pulling together positive pairs while pushing apart the negatives. To alleviate the influence of intrinsic false-negative pairs and rectify cluster assignments, we adopt a confidence-based criterion to select pseudo-labels for boosting both the instance- and cluster-level contrastive learning. As a result, the clustering performance is further improved. Besides the elegant idea of twin contrastive learning, another advantage of TCL is that it could independently predict the cluster assignment for each instance, thus effortlessly fitting online scenarios. Extensive experiments on six widely-used image and text benchmarks demonstrate the effectiveness of TCL. The code will be released on GitHub.
\end{abstract}

\begin{figure}[t]\centering
    \includegraphics[width=0.8\columnwidth]{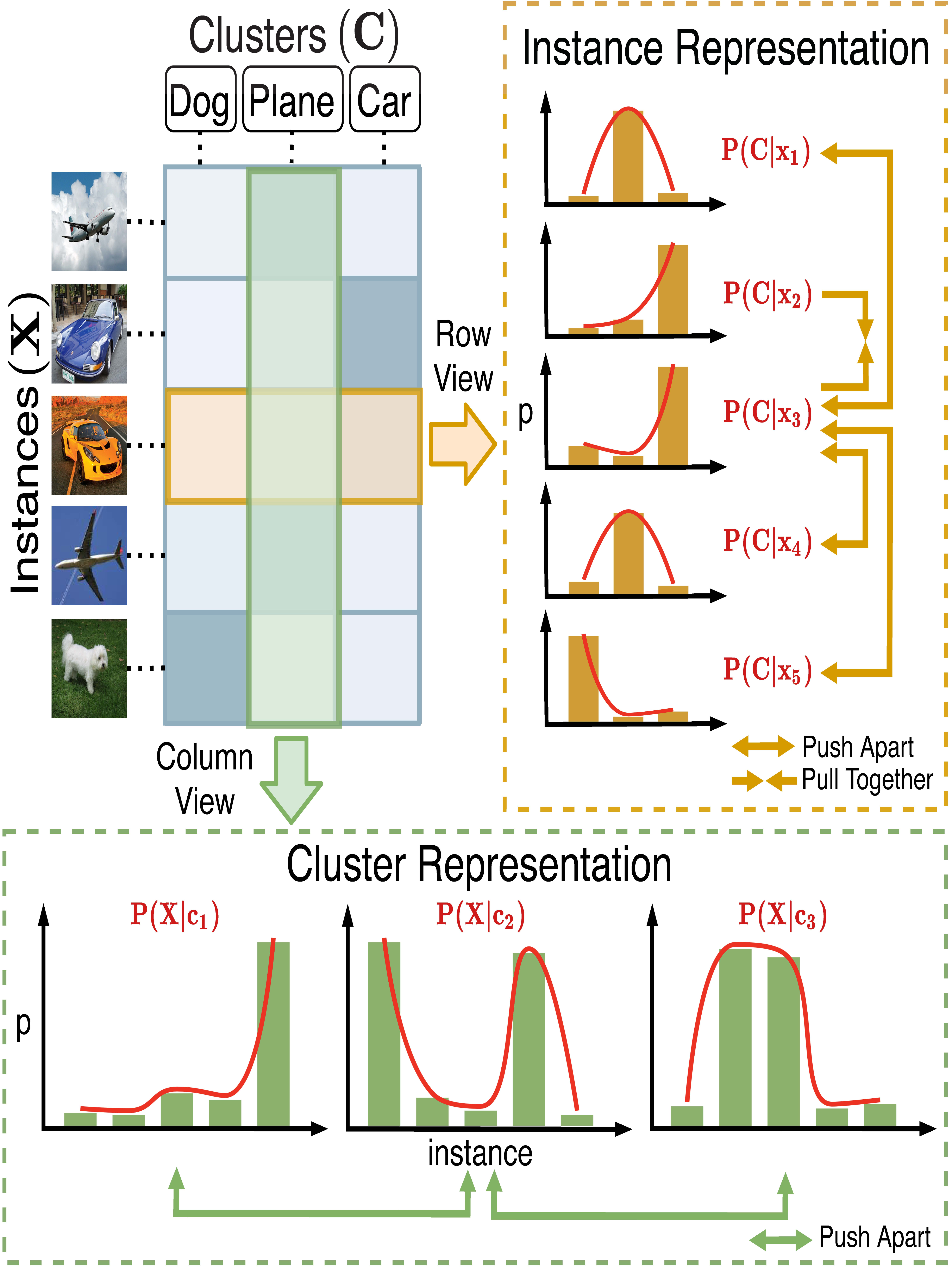}
    \caption{Our key observation and basic idea. By projecting data into a feature space with the dimensionality of cluster number, the element in the $i$-th row and $k$-th column of the feature matrix represents the probability of instance $i$ belonging to cluster $k$. Namely, rows correspond to the cluster assignment probabilities, which are special representations of instances. More interestingly, if we look at the feature matrix from the column view, each column actually corresponds to the cluster distribution over the data, which could be seen as a special representation of the cluster. As a result, the instance- and cluster-level representation learning (\textit{e.g.}, contrastive learning) could be conducted in the row and column space, respectively.}
    \label{fig:overview}
\end{figure}

\section{Introduction}\label{sec:introduction}

Clustering is one of the most fundamental tasks in machine learning and data mining. It aims to group data into different clusters without label information, such that the within-cluster data come from the same class or share similar semantics. Besides facilitating general representation learning~\citep{DeepClustering, SWAV}, clustering is also helpful in a variety of real-world applications, such as face recognition~\citep{FaceCluster}, medical analysis~\citep{Medical}, and gene sequencing~\citep{RNA}.

During past years, most clustering methods~\citep{chen2009spectral, nie2011spectral, GDL, liu2016multiple, liu2017sparse, nie2019k, tang2019kernel} mainly focus on developing different similarity metrics and clustering strategies. Though grounded in theory, their performance is limited by the adopted shallow models. Recently, deep clustering~\citep{ghasedi2017deep, deepclustering_wangqi} has shown promising results on various benchmarks by extracting representative features to facilitate downstream clustering. Early deep clustering methods~\citep{peng2016deep, JULE, DeepClustering, DEC} iteratively perform representation learning and clustering to bootstrap each other. However, this kind of method usually needs the entire dataset to perform offline clustering, which is less attractive for large-scale data and even impractical for streaming data. Luckily, the offline limitation could be solved by the idea of ``label as representation''~\citep{Peng2017:Cascade_full, LabelAsRep}. By directly and independently predicting cluster assignment for each instance, large-scale and online clustering could be achieved~\citep{IMSAT, IIC, PICA}. Very recently, the rapid growth of contrastive learning~\citep{SimCLR, BarlowTwins} significantly improves the performance of unsupervised representation learning. Motivated by their successes, some contrastive learning based clustering methods~\citep{SCAN, CC, SPICE, han2020mitigating} are proposed, which achieve state-of-the-art results.

In this work, we propose an end-to-end online deep clustering method by conducting twin contrastive learning (TCL) based on the observation shown in Fig.~\ref{fig:overview}. In brief, the rows and columns of the feature matrix correspond to the instance and cluster representations, respectively. Under this observation, TCL conducts contrastive learning in the row and column space of the feature matrix to jointly learn the instance and cluster representation. Specifically, TCL first constructs contrastive pairs through data augmentations. Different from most existing contrastive learning methods that use weak augmentations proposed in SimCLR~\citep{SimCLR}, we provide a new effective augmentation strategy by mixing weak and strong transformations. With the constructed pairs, TCL performs contrastive learning at both the instance and cluster level. The instance-level contrastive learning aims to pull within-class instances together while pushing between-class instances apart. And the cluster-level contrastive learning aims to distinguish distributions of different clusters while attracting distributions of the same cluster under different augmentations. To relieve the influence of intrinsic false-negative pairs and rectify cluster assignments, we progressively select the confident predictions (\textit{i.e.}, those with cluster assignment probability close to one-hot) to fine-tune the twin contrastive learning. Such a fine-tuning strategy is based on the observation that the predictions with high confidence are more likely to be correct and thus could be used as pseudo labels. Once the model converges, it could independently make cluster assignments for each instance in an end-to-end manner to achieve clustering. The major contributions of this work are summarized as follows:
 \begin{itemize}
 	\item We reveal that the rows and columns of the feature matrix intrinsically correspond to the instance and cluster representations. On top of that, we propose TCL that achieves clustering by simultaneously conducting contrastive learning at the instance and cluster level;
 	\item We provide a new data augmentation strategy by mixing weak and strong transformations, which naturally fits our TCL framework and is proved to be effective for both image and text data in our experiments;
 	\item To alleviate the influence of intrinsic false negatives and rectify cluster assignments, we adopt a confidence-based criterion to generate pseudo-labels for fine-tuning both the instance- and cluster-level contrastive learning. Experiments show that such a fine-tuning strategy could further boost the clustering performance;
 	\item The proposed TCL clusters data in an end-to-end and online manner, which only needs batch-wise optimization and thus could handle large-scale datasets. Moreover, TCL could handle streaming data since it could timely make cluster assignments for new coming data without accessing the whole dataset.
\end{itemize}

\section{Related Work}

In this section, we give a brief review on contrastive learning and deep clustering, followed by a discussion on the connection between these two topics.

\subsection{Contrastive Learning}

Recently, the contrastive learning paradigm shows its power in unsupervised representation learning~\citep{MOCO, SimCLR, BYOL, SimSiam, AdCo, SWAV, BarlowTwins}. It first constructs positive and negative pairs for each instance and then projects them into a subspace to maximize the similarities of positive pairs and minimize those of the negatives~\citep{DimensionalityReductionbyLearninganInvariantMapping}. The most straightforward solution is to use labels to guide the pair construction~\citep{SupervisedCL}. However, in the unsupervised setting, other strategies are needed to construct and utilize contrastive pairs. For example, SimCLR~\citep{SimCLR} constructs positive and negative pairs through augmentations within mini-batch. MoCo~\citep{MOCO} recasts contrastive learning as a dictionary look-up task by building a dynamic dictionary with a queue and a moving-averaged encoder. To avoid the efforts in building negative pairs, BYOL~\citep{BYOL} and SimSiam~\citep{SimSiam} replace negative pairs with an online predictor that prevents the network from collapsing into trivial solutions. As an alternative, AdCo~\citep{AdCo} directly learns negative samples in an adversarial manner. Lately, Barlow Twins~\citep{BarlowTwins} performs contrastive learning from a redundancy-reduction perspective and achieves comparable results.

There are two major differences between our work and these contrastive learning methods. First, our method concurrently conducts row- and column-wise contrastive learning at both the instance and cluster level while most existing methods solely perform row-wise contrastive learning at the instance level. Such an elegant idea is based on our observation that rows and columns of the feature matrix correspond to the instance and cluster representations respectively. Second, the aforementioned methods adopt the weak augmentation strategy proposed in SimCLR~\citep{SimCLR} because strong augmentations (RandAugment~\citep{RandAug} to be specific) experimentally show inferior performance~\citep{StrongContrastive}. Though there are some works~\citep{SCAN} that use strong augmentations to fine-tune the network, it is still unclear how to directly facilitate contrastive learning with strong augmentations. From this perspective, this work could shed some light on how to effectively utilize weak and strong transformations by using the proposed TCL framework (more details in Table~\ref{tab:gain}). The proposed augmentation strategy is suitable for various types of data, such as images and texts.

\begin{figure*}[t]\centering
    \includegraphics[width=1.8\columnwidth]{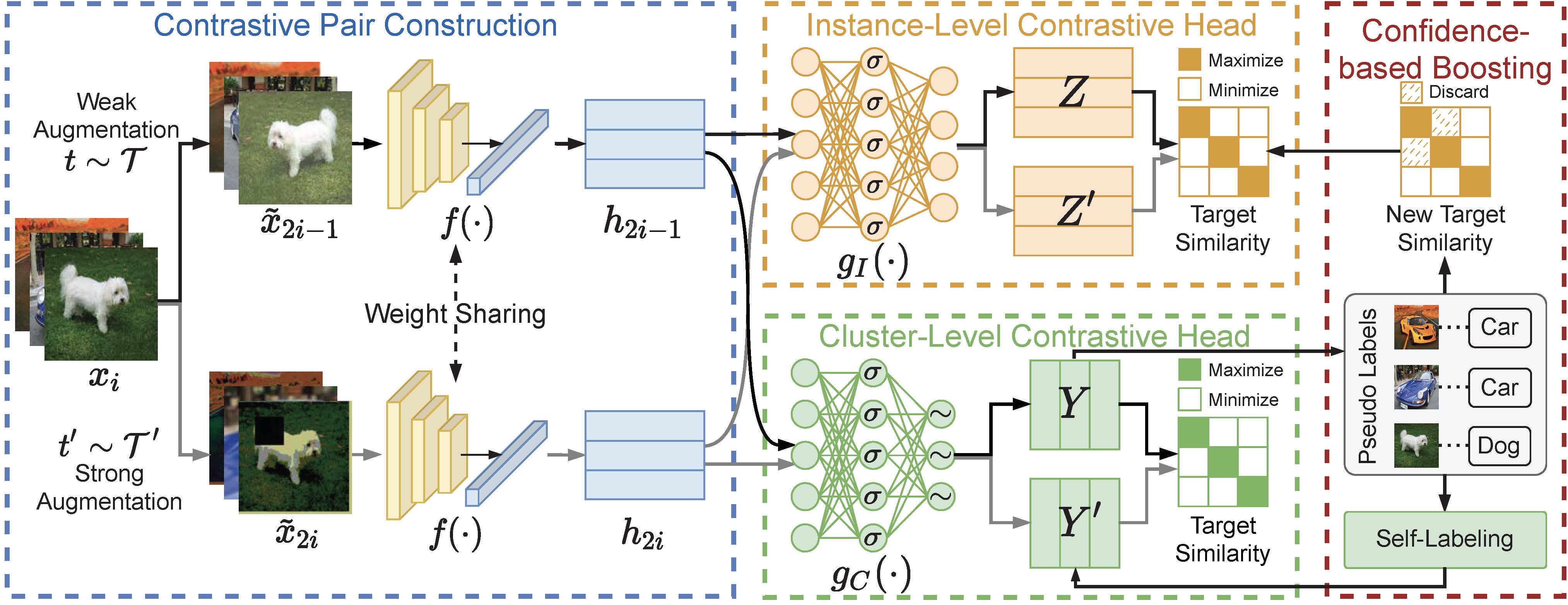}
    \caption{The pipeline of Twin Contrastive Learning (TCL). First, it constructs data pairs through weak and strong augmentations. A shared backbone is used to extract features from augmented samples. Then, two independent MLPs ($\sigma$ denotes the ReLU activation and $\sim$ denotes the Softmax operation to produce soft labels) project the features into the row and column space wherein the instance- and cluster-level contrastive learning are conducted, respectively. Finally, pseudo labels are selected based on the confidence of cluster predictions to alleviate the influence of false-negative pairs and rectify previous predictions, which further boosts the clustering performance.}
    \label{fig:framework}
\end{figure*}

\subsection{Deep Clustering}
Both effective clustering strategies and discriminative features are essential in achieving good clustering. Thanks to the powerful representability of deep neural networks, deep clustering methods have attracted more and more attention in recent~\citep{peng2016deep, DEC, guo2017improved, deepclustering_wangqi, peng2018structured, peng2022xai, JULE, DeepClustering, SelfLabling, li2020prototypical}. To name a few, JULE~\citep{JULE} iteratively learns data representation and performs hierarchical clustering. DeepCluster~\citep{DeepClustering} clusters data using the prior representation and uses the cluster assignment of each sample as a classification target to learn the new representation. Though representation learning and clustering could bootstrap each other to some extent, this kind of two-stage method might suffer from errors accumulated during alternations. Another weakness of these methods is that they could not be applied in the online scenario, where data is presented in streams and only a batch of samples are accessible at one time. Specifically, JULE needs the global similarity to decide which sub-clusters should be merged, while DeepCluster and SL~\citep{SelfLabling} need to perform offline k-means or solve a global optimal transport problem to acquire cluster assignments. To overcome the offline limitation, some online deep clustering methods are proposed~\citep{CC, zhong2020deep, dang2021doubly, IIC, PICA}. For example, IIC~\citep{IIC} discovers clusters by maximizing mutual information between the cluster assignments of data pairs. PICA~\citep{PICA} learns the most semantically plausible data separation by maximizing the partition confidence of the clustering solution. Very recently, some studies~\citep{SCAN, RUC, SPICE} use pseudo-labels generated by preliminary clustering, namely, self-labeling, to further improve the clustering performance in a multi-stage manner.

Unlike most of the above works that perform representation learning and clustering in multiple stages, our method unifies these two tasks into the twin contrastive learning framework. Such a one-stage learning paradigm helps the model to learn more clustering-favorable representations compared with previous works that solely conduct instance-level contrastive learning~\citep{SCAN, SPICE}. In the boosting stage, despite rectifying the cluster assignments based on the features extracted in the early stage~\citep{SPICE}, we could also fine-tune the instance-level contrastive learning to alleviate the influence of false-negative pairs thanks to our one-stage learning paradigm.

\subsection{Connection between contrastive learning and deep clustering}

Both representation learning and deep clustering share a common goal, namely, extracting discriminative features. Recently, a variety of works have shown that contrastive learning and deep clustering could bootstrap each other. On the one hand, the performance of representation learning could be enhanced by integrating the clustering property~\citep{wang2021unsupervised}. For example, instead of constructing pairs between samples, SwAV~\citep{SWAV} first performs clustering by solving an optimal transport problem and then contrasts the instances and the learned cluster centers. Similarly, PCL~\citep{li2020prototypical} pulls each sample to its corresponding cluster center with a prototypical contrastive loss. In addition, \cite{van2021revisiting} and \cite{dwibedi2021little} show that leveraging the local neighborhood can be effective for contrastive learning. On the other hand, recent deep clustering methods such as SCAN~\citep{SCAN}, CC~\citep{CC}, and SPICE~\citep{SPICE} have achieved state-of-the-art performance thanks to the contrastive learning paradigm.

Though the combination of contrastive learning and deep clustering has brought promising results, most existing works still treat these two tasks separately. Different from existing works, this study elegantly unifies contrastive learning and deep clustering into the twin contrastive learning framework, which might bring some insights to both communities. Notably, this study is a significant extension of \citep{CC} with the following improvements:
\begin{itemize}
	\item In this paper, we propose a confidence-based boosting strategy to fine-tune both the instance- and cluster-level contrastive learning. Specifically, most confident predictions are selected as pseudo labels based on the observation that they are more likely to be correct. Upon that, we use the generated pseudo labels to alleviate the influence of false-negative pairs (composed of within-class samples) in instance-level contrastive learning, and adopt cross-entropy loss to rectify cluster assignments in cluster-level contrastive learning. Notably, such a twin self-training paradigm is benefited from our TCL framework since the cluster assignments (from CCH) of instance features (from ICH) could be obtained in an online manner.
	\item In this paper, we propose a data augmentation strategy by mixing weak and strong transformations. Though such an augmentation strategy is seemingly simple, its effectiveness is closely correlated with the proposed TCL framework. Previous works have shown that directly introducing strong augmentation into the contrastive learning framework could lead to sub-optimal performance~\citep{StrongContrastive}. Different from such a conclusion, we show that the mixed augmentation strategy naturally fits the proposed TCL framework (see Table~\ref{tab:gain} for more details).
	\item To investigate the generalization ability of the proposed method, we verify the effectiveness of our method in text clustering despite the difference in data augmentation. Experimental results demonstrate the superiority of the proposed TCL framework, mixed augmentation strategy, and confidence-based boosting strategy. A comparable performance gain is achieved by this journal extension compared with the previous conference paper~\citep{CC}.
\end{itemize}

\section{Method}

The pipeline of the proposed TCL is illustrated in Figure~\ref{fig:framework}. The model consists of three parts, namely, the contrastive pair construction (CPC), the instance-level contrastive head (ICH), and the cluster-level contrastive head (CCH), which are jointly optimized through twin contrastive learning and confidence-based boosting. Specifically, in the twin contrastive learning stage, CPC first constructs contrastive pairs through data augmentations and then projects the pairs into a latent feature space. After that, ICH and CCH conduct instance- and cluster-level contrastive learning at the row and column space of the feature matrix respectively by minimizing the proposed twin contrastive loss. To alleviate the influence of intrinsic false-negative pairs in contrastive learning and to rectify cluster assignments, we propose a confidence-based boosting strategy (CB). In detail, some confident predictions are selected as pseudo labels to fine-tune the instance- and cluster-level contrastive learning with the self-supervised contrastive loss and self-labeling loss, which further improves the clustering performance.

Once the model converges, CCH could make cluster assignments for each instance to achieve online clustering. Notably, though the twin contrastive learning can be directly conducted on the same contrastive head as indicated in our basic idea, we experimentally find that decoupling it into two independent subspaces improves the clustering performance (see Section~\ref{sec:decouple} for detailed discussion).

In this section, we first introduce the construction of contrastive pairs in CPC, then present the twin contrastive loss for training, and finally elaborate on our confidence-based boosting strategy.

\subsection{Contrastive Pair Construction}

Inspired by recent developments of contrastive learning~\citep{SimCLR, SWAV}, the proposed TCL constructs contrastive pairs through data augmentation. Specifically, for each instance $x_i$, CPC stochastically samples and applies two groups of transformations $t$ and $t^\prime$ from two augmentations families $\mathcal{T}$ and $\mathcal{T}^\prime$ respectively, resulting in two correlated samples (\textit{i.e.}, a data pair) denoted as $\tilde{x}_{2i-1} = t(x_i)$ and $\tilde{x}_{2i} = t^\prime(x_i)$.

Recent studies suggest that data augmentation is essential in contrastive learning methods~\citep{SimCLR, SWAV}, and most of them adopt weak augmentations~\citep{SimCLR, BYOL, BarlowTwins} since directly using strong augmentations experimentally leads to inferior performance~\citep{StrongContrastive}. In this work, we provide a novel data augmentation strategy by mixing weak and strong transformations, which achieves superior performance on both image and text data. To be specific, for image data, we adopt the transformations proposed by SimCLR~\citep{SimCLR} and RandAugment~\citep{RandAug} as the weak $\mathcal{T}$ and strong $\mathcal{T}^\prime$ augmentation, respectively. For text data, we employ the synonym replacement strategy~\citep{SCCL} as the weak augmentation $\mathcal{T}$ and use the sentence operations~\citep{EDA} as the strong augmentation $\mathcal{T}^\prime$.

Given the constructed pairs, a shared backbone $f(\cdot)$ is used to extract features $h$ from the augmented samples through $h_{2i-1}=f(\tilde{x}_{2i-1})$ and $h_{2i}=f(\tilde{x}_{2i})$. Specific backbones are used to handle different types of data. In this work, we adopt ResNet~\citep{ResNet} and Sentence Transformer~\citep{SentenceTransformer} as the backbone for image and text data, respectively.

\subsection{Twin Contrastive Learning}

In the training stage, the backbone, the instance-level contrastive head (ICH), and the cluster-level contrastive head (CCH) are jointly optimized according to the following twin contrastive loss, \textit{i.e.},
\begin{equation}
\label{eq:overall loss}
  \mathcal{L}_{train} = \mathcal{L}_{ins} + \mathcal{L}_{clu},
\end{equation}
where $\mathcal{L}_{ins}$ is the instance-level contrastive loss which is computed on ICH and $\mathcal{L}_{clu}$ denotes the cluster-level contrastive loss computed on CCH.

In general, one may add a dynamic weight parameter to balance the two losses across the training process, but explicitly tuning the weight could violate the unsupervised constraint. In practice, we find a simple addition of the two contrastive losses already works well.

\subsubsection{Instance-level contrastive loss}

The instance-level contrastive learning aims to maximize the similarities of positive pairs while minimizing those of negative ones. To achieve clustering, ideally, one could define pairs of within-class instances to be positive and those of between-class instances to be negative. However, since no prior label information is given, we construct instance pairs based on data augmentations as a compromise. To be specific, the positive pairs consist of samples augmented from the same instance, and the negative pairs otherwise.

Formally, for a mini-batch of size $N$, TCL performs two types of data augmentations on each instance $x_i$, resulting in $2N$ augmented samples $\{\tilde{x}_{1},\tilde{x}_{2},\dots,\tilde{x}_{2i-1},\tilde{x}_{2i},$ $\dots,\tilde{x}_{2N}\}$. Each sample $\tilde{x}_{2i-1}$ forms $2N-1$ pairs with others, among which we choose the pair with its corresponding augmented sample $\{\tilde{x}_{2i-1}, \tilde{x}_{2i}\}$ to be positive and define other $2N-2$ pairs to be negative.

As directly conducting contrastive learning on the feature matrix may cause information loss~\citep{SimCLR}, we stack a two-layer nonlinear MLP $g_I(\cdot)$ to map the features into a subspace via $z_i=g_I(h_i)$, where the instance-level contrastive learning is applied. The pair-wise similarity is measured using cosine distance, namely,
\begin{equation}
\label{eq:instance similarity}
  s(z_i, z_j) = \frac{z_i z_j^\top}{\|z_i\|\|z_j\|}, i,j \in [1, 2N].
\end{equation}

The InfoNCE loss~\citep{CPC} is adopted to optimize pair-wise similarities defined by Eq.~\ref{eq:instance similarity}. Without loss of generality, the loss for a given augmented sample $\tilde{x}_{i}$ (suppose it forms a positive pair with $\tilde{x}_{j}$) is defined as
\begin{equation}
  \ell_{i} = -\log \frac{\exp(s(z_{i}, z_{j})/\tau_I)}
  {\sum_{k=1}^{2N} \mathbbm{1}_{[k \neq i]} \exp \left(s({z}_{i}, {z}_{k}) / \tau_I\right)},
\end{equation}
where $\tau_I$ is the instance-level temperature parameter to control the softness, and $\mathbbm{1}_{[k \neq i]}$ is an indicator function evaluating to $1$ iff $k \neq i$. To identify the positive counterpart for each augmented sample, the instance-level contrastive loss is computed across all augmented samples, \textit{i.e.},
\begin{equation}
\label{eq:instance loss}
  \mathcal{L}_{ins} = \frac{1}{2N} \sum_{k=1}^{2N} \ell_{k}.
\end{equation}

\subsubsection{Cluster-level contrastive loss}

When a sample is projected into a subspace whose dimensionality equals the cluster number, the $i$-th element of its feature represents its probability of belonging to the $i$-th cluster. In other words, the feature vector corresponds to its cluster assignment probability.

Suppose the target cluster number is $M$, similar to the instance-level contrastive head, we use another two-layer MLP $g_C(\cdot)$ to project the features into an $M$-dimensional space via $y_i=g_C(h_i)$. Here $y_i$ corresponds to the cluster assignment probability of the augmented sample $\tilde{x}_i$. Formally, let $Y=[y_1, \dots, y_{2i-1}, \dots, y_{2N-1}] \in \mathcal{R}^{N\times M}$ be the cluster assignment probabilities of the mini-batch under the weak augmentation $\mathcal{T}$ (and $Y^\prime=[y_2, \dots, y_{2i}, \dots, y_{2N}]$ for those under the strong augmentation $\mathcal{T}^\prime$). Based on the observation shown in Fig.~\ref{fig:overview}, the columns of $Y$ and $Y^\prime$ correspond to the cluster distributions over the mini-batch and could be interpreted as special cluster representations. We would like to point out that this observation still holds even when the dimension is larger than the ground-truth cluster number. In that case, a more fine-grained cluster structure is considered and its effectiveness is verified in Barlow Twins~\citep{BarlowTwins}.

For clarity, we denote the $i$-the column of $Y$ as $\hat{y}_{2i-1}$ (and $\hat{y}_{2i}$ for the $i$-the column of $Y^\prime$), namely, the representation of cluster $i$ under the weak (and strong) data augmentation. The representations of the same cluster under two augmentations form positive cluster pairs $\{\hat{y}_{2i-1}, \hat{y}_{2i}\}, i\in [1,M]$, while other pairs are defined to be negative. Again, we use cosine distance to measure the similarity between cluster $\hat{y}_{i}$ and $\hat{y}_{j}$, that is
\begin{equation}
\label{eq:cluster similarity}
  s(\hat{y}_i, \hat{y}_j) = \frac{\hat{y}_i^\top\hat{y}_j}{\|\hat{y}_i\|\|\hat{y}_j\|}, i, j \in [1, 2M]
\end{equation}

Without loss of generality, the following loss function is adopted to identify cluster $\hat{y}_i$ from all other $2M-2$ clusters except its counter part $\hat{y}_j$, \textit{i.e.},
\begin{equation}
  \hat\ell_{i} = -\log \frac{\exp(s(\hat{y}_i, \hat{y}_j)/\tau_C)}
  {\sum_{k=1}^{2M} \mathbbm{1}_{[k \neq i]} \exp \left(s(\hat{y}_i, \hat{y}_k) / \tau_C\right)},
\end{equation}
where $\tau_C$ is the cluster-level temperature parameter to control the softness, and $\mathbbm{1}_{[k \neq i]}$ is an indicator function evaluating to $1$ iff $k \neq i$. By traversing all clusters, the cluster-level contrastive loss is computed through
\begin{equation}
  \mathcal{L}^\prime_{clu} = \frac{1}{2M} \sum_{k=1}^M \hat{\ell}_{k}.
\end{equation}

As simply optimizing the above cluster-level contrastive loss might lead to trivial solution where most samples are assigned to a few clusters, we add a cluster entropy to prevent the model from collapsing and achieve more balanced clustering~\citep{ghasedi2017deep, PICA}. Formally, let $P(\hat{y}_{2i-1}) = \frac{1}{N}\sum_{k=1}^N Y_{ki}$ be the assignment probability of cluster $i$ within a mini-batch under the weak augmentation and $P(\hat{y}_{2i}) = \frac{1}{N}\sum_{k=1}^N Y^\prime_{ki}$ be that under the strong augmentation, then the cluster entropy is computed by
\begin{equation}
	H_{clu}=-\sum_{i=1}^{2M} [P(\hat{y}_{i})\log P(\hat{y}_{i})].
\end{equation}

To sum up, the cluster-level contrastive loss is finally defined as
\begin{equation}
\label{eq:cluster loss}
  \mathcal{L}_{clu} = \frac{1}{2M} \sum_{k=1}^{2M} \hat{\ell}_{k}-H_{clu}.
\end{equation}

\begin{algorithm}[!t]
\caption{Twin Contrastive Learning}
\label{algorithm}
\KwIn{dataset $\mathcal{X}$; training iterations $E_1$; boosting iterations $E_2$; batch size $N$; cluster number $M$; temperature parameter $\tau_I$ and $\tau_C$; network $f$, $g_I$, and $g_C$; augmentation strategies $\mathcal{T},\mathcal{T}^\prime$.}

\KwOut{cluster assignments.}

\tcp{Training}

\For{epoch = 1 to $E_1$}{
	sample a mini-batch $\{x_i\}_{i=1}^N$ from $\mathcal{X}$
		
	sample two augmentations $t\sim \mathcal{T}, t^\prime \sim \mathcal{T}^\prime$
	
	compute instance and cluster representations
	
	compute instance-level contrastive loss $\mathcal{L}_{ins}$ through Eq.~\ref{eq:instance similarity}--\ref{eq:instance loss}
	
	compute cluster-level contrastive loss $\mathcal{L}_{clu}$ through Eq.~\ref{eq:cluster similarity}--\ref{eq:cluster loss}
	
	compute training loss $\mathcal{L}_{train}$ by Eq.~\ref{eq:overall loss}
	
	update $f, g_I, g_C$ through gradient descent to minimize $\mathcal{L}_{train}$
}

\tcp{Boosting}

\For{epoch = 1 to $E_2$}{
	sample a mini-batch $\{x_i\}_{i=1}^N$ from $\mathcal{X}$
	
	sample two augmentations $t\sim \mathcal{T}, t^\prime \sim \mathcal{T}^\prime$
	
	update pseudo labels with Eq.~\ref{eq:conf&pred}--\ref{eq:pseudolabel} and Eq.~\ref{eq:weedout}
	
	compute self-supervised contrastive loss $\mathcal{L}_{scl}$ by Eq.~\ref{eq:scl}
	
	compute self-labeling loss $\mathcal{L}_{sl}$ by Eq.~\ref{eq:sl}
	
	compute boosting loss $\mathcal{L}_{boost}$ by Eq.~\ref{eq:boost_loss}
	
	update $f, g_I, g_C$ through gradient descent to minimize $\mathcal{L}_{boost}$
}

\tcp{Test}

\For{$x$ in $\mathcal{X}$}{
	extract features by $h = f(x)$
	
	compute cluster assignment by $c=\arg\max g_C(h)$
}
\end{algorithm}

\subsection{Confidence-based Boosting}

As the train progresses, we notice that the model tends to make more confident predictions (\textit{i.e.}, with cluster assignment probability close to one-hot). Those confident predictions are more likely to be correct (see Fig.~\ref{fig:performance}). Based on this observation, in the boosting stage, we progressively select the most confident predictions as pseudo labels to fine-tune both the instance- and cluster-level contrastive learning. The pseudo labels are selected by the following criterion. Namely, for a mini-batch of size $N$, we use the raw data $x$ as input to compute prediction $\mathrm{pred}$ with confidence $conf$ for each instance by
\begin{equation}
\label{eq:conf&pred}
\begin{aligned}
	y_i &= g_C(f(x_i)),\\
	\mathrm{conf}_i &= \max(y_i), \\
	\mathrm{pred}_i&=\arg \max(y_i).
\end{aligned}
\end{equation}

In every mini-batch, we select the top $\gamma$ confident predictions from each cluster as pseudo labels, where $\gamma$ is the confident ratio and we fix it to $0.5$. To be specific, a prediction $\mathrm{pred}_i$ will be selected as pseudo label if it meets the following criteria, \textit{i.e.},
\begin{equation}
\label{eq:pseudolabel}
\begin{aligned}
	n &= \gamma \times N / M, \\
	\mathrm{CONF}_k &= sort(\{\mathrm{conf}_i \mid i\in [1,N], \mathrm{pred}_i=k\})[n],\\
	\mathrm{conf}_i &\geq \mathrm{CONF}_{\mathrm{pred}_i},
\end{aligned}
\end{equation}
where $\mathrm{CONF}_k$ is the $n$-th largest confidence of predictions on cluster $k\in[0,M-1]$. Notably, selecting most confident predictions from each cluster leads to more class-balanced pseudo labels compared with threshold-based criterion~\citep{SCAN, RUC, SPICE}. We store the pseudo labels for all instances, denoted as $P$, in the memory.

With the generated pseudo labels, we fine-tune the model with the following loss to further boost the clustering performance, namely,
\begin{equation}
	\mathcal{L}_{boost} =\mathcal{L}_{scl}+\mathcal{L}_{sl},
\label{eq:boost_loss}
\end{equation}
where $\mathcal{L}_{scl}$ is the self-supervised contrastive loss used to alleviate the influence of false negative pairs in instance-level contrastive learning and $\mathcal{L}_{sl}$ is the self-labeling loss for rectifying cluster assignments made by CCH. 

Recall that in the instance-level contrastive learning, we treat pairs of samples augmented from different instances to be negative, since no label information is given. However, for downstream tasks such as classification and clustering, within-class samples should not be pushed apart. To this end, with the help of pseudo labels, we remove within-class samples from negatives pairs (\textit{i.e.}, the denominator in Eq.~\ref{eq:instance loss}) and adopt a self-supervised contrastive loss to fine-tune the instance-level contrastive learning. Specifically, for each augmented sample $x_i$ with pseudo label $\mathrm{pred}_i$, the self-supervised contrastive loss is defined as
\begin{equation}
	l^\prime_{i}=-\log \frac{\exp \left(s(z_{i},z_{j}) / \tau_I \right)}{\sum_{k=1}^{2 N} \mathbbm{1}_{[\mathrm{pred}_{k}\neq\mathrm{pred}_{i}]} \cdot \exp \left(s(z_{i},z_{k}) / \tau_I \right)}, \\
\end{equation}
where $\mathbbm{1}$ is the indicator. Notably, inspired by the negative learning paradigm~\citep{kim2019nlnl}, here we only remove potential within-class pairs from negative ones without treating them as positive, considering that the latter strategy could be too strong when the pseudo labels are of inferior quality. By traversing all augmented samples, the self-supervised instance-level contrastive loss is computed through
\begin{equation}
\label{eq:scl}
	\mathcal{L}_{scl}=\frac{1}{2N} \sum_{i=1}^{2 N} l^\prime_{i}.
\end{equation}

For the cluster-level contrastive head, the self-labeling strategy is adopted to rectify previous predictions. Specifically, we define the self-labeling loss as the weighted cross-entropy on the strongly augmented samples $x^\prime=t^\prime (x)$, \textit{i.e.},
\begin{equation}
\label{eq:sl}
\begin{aligned}
y^\prime_i&=g_C(f(x^\prime_i)),\\
\mathcal{L}_{sl}&=-\frac{1}{N_p}\sum_{i=1, i\in P}^{N} w_{\mathrm{pred}_i}\log\left(\frac{\exp(y^\prime_i[\mathrm{pred}_i])}{\sum_{k=0}^{M-1} \exp(y^\prime_i[k])}\right),
\end{aligned}
\end{equation}
where $N_p$ denotes the number of instances that have pseudo-labels in the mini-batch, and $w_c\propto{1}/{N_c}$ is the weight parameter for cluster $c$ of size $N_c$. The weighted loss could prevent large clusters from dominating the optimization.

Though the confident ratio is fixed to $\gamma=0.5$, it does not mean that at most $50\%$ of predictions will be selected as pseudo labels. As the boosting progresses and batch shuffles, more pseudo labels would be selected progressively. Furthermore, considering that the model might make some mistakes in selecting pseudo labels, we remove the pseudo labels from $P$ once their confidences decrease to below a certain threshold, namely, 
\begin{equation}
\label{eq:weedout}
	\mathrm{conf}_i<\alpha,
\end{equation}
where $\alpha$ is the lower confidence bound of pseudo labels and is set to $0.99$. This weeding out mechanism keeps the high quality of pseudo labels and gives the model a chance of rectifying previous predictions. The choice of $\alpha$ and $\gamma$ is fixed across all the experiments, and we provide parameter analysis on them in Section~\ref{sec:hyper-params}.

The overall training, boosting, and test process of the proposed TCL is summarized in Algorithm~\ref{algorithm}.

\section{Experiments}

In this section, the clustering performance of the proposed TCL is evaluated on five image and two text datasets. A series of qualitative analyses and ablation studies are carried out to help comprehensively and intuitively understand the method.

\subsection{Datasets}

A brief description of the used datasets is summarized in Table~\ref{tab:datasets}. More specifically, the image datasets include CIFAR-10, CIFAR-100~\citep{CIFAR}, STL-10~\citep{STL}, ImageNet-10, and ImageNet-Dogs~\citep{ImageNet10/dogs}. For CIFAR-100,  its 20 super-classes rather than 100 fine-grained classes are taken as the ground truth. The text datasets include StackOverflow~\citep{TextData} and Biomedical~\citep{TextData}. The former is a subset of the challenge data published by Kaggle, and the latter is a subset of the PubMed data distributed by BioASQ.

\begin{table}[h]
\centering
\caption{A summary of datasets used for evaluation.}
\begin{tabular}{@{}cccc@{}}
\toprule
Dataset       & Split      & Samples & Classes \\ \midrule
CIFAR-10      & Train+Test & 60,000   & 10      \\
CIFAR-100     & Train+Test & 60,000   & 20      \\
STL-10        & Train+Test & 13,000   & 10      \\
ImageNet-10   & Train      & 13,000   & 10      \\
ImageNet-Dogs & Train      & 19,500   & 15      \\ \midrule
StackOverflow & -          & 20,000    & 20      \\
Biomedical    & -          & 20,000    & 20      \\ \bottomrule
\end{tabular}
\label{tab:datasets}
\end{table}

\subsection{Experimental Settings}

Different backbones could be used to handle different types of data, including but not limited to images and texts. For a fair comparison with previous works~\citep{IIC, PICA, TextCluster}, we adopt ResNet34~\citep{ResNet} as the backbone for images and the \textit{distilbert-base-nli-stsb-mean-tokens} model from the Sentence Transformer library~\citep{SentenceTransformer} for texts. For ResNet34, its output dimension of the fully-connected classification layer (\textit{i.e.}, the dimension of $h$) is set to 512. For STL-10, its 100,000 unlabeled images are additionally used to compute the instance-level contrastive loss in the training stage. The ResNet34 is randomly initialized, while the Sentence Transformer is pre-trained to produce meaningful word embeddings to keep consistent with previous works~\citep{TextData, TextCluster}. For simplicity, instead of customizing the network to handle images of different sizes, we simply resize all images to $224 \times 224$, and no additional modification is made on the standard ResNet34.

For images, we adopt data transformations proposed in SimCLR~\citep{SimCLR} as weak augmentation, including ResizedCrop, ColorJitter, Grayscale, HorizontalFlip, and GaussianBlur. Notably, as small images already become blurred after up-scaling, we leave the GaussianBlur augmentation out for CIFAR-10/100 in the weak augmentation $\mathcal{T}$. The strong augmentation $\mathcal{T}^\prime$ is composed of four randomly selected transformations from RandAugment~\citep{RandAug} with parameters listed in Table~\ref{tab:strong_aug_image}, followed by one Cutout~\citep{CutOut} operation with a size of $75\times 75$.

\begin{table}[h]
\centering
\caption{List of strong augmentations for images.}
\begin{tabular}{@{}ccc@{}}
\toprule
Transformations       & Parameter      & Range \\ \midrule
AutoContrast      & - & -  \\
Equalize     & - & -  \\
Identity        & - & -  \\
Brightness   & B      & [0.05, 0.95]  \\
Color & C      & [0.05, 0.95]  \\
Contrast & C      & [0.05, 0.95]  \\
Posterize & B      & [4, 8]  \\
Rotate & $\theta$      & [-30, 30]  \\
Sharpness & S      & [0.05, 0.95]  \\
Shear X, Y & R      & [-0.3, 0.3]  \\
Solarize & T      & [0, 256] \\
Translate X, Y &   $\lambda$         & [-0.3, 0.3]  \\ \bottomrule
\end{tabular}
\label{tab:strong_aug_image}
\end{table}

\begin{table*}[t]
\centering
\caption{The clustering performance on five object image benchmarks. The first and second best results are shown in \textbf{bold} and \underline{underline}, respectively. ``()'' denotes that extra training data is used. ``TCL$_\text{ICH}$'' refers to clustering results achieved by conducting $k$-means on the ICH features. }
\resizebox{\textwidth}{!}{%
\begin{tabular}{@{}lccccccccccccccc@{}}
\toprule
Dataset &
  \multicolumn{3}{c}{CIFAR-10} &
  \multicolumn{3}{c}{CIFAR-100} &
  \multicolumn{3}{c}{STL-10} &
  \multicolumn{3}{c}{ImageNet-10} &
  \multicolumn{3}{c}{ImageNet-Dogs} \\ \midrule
Metrics &
  \multicolumn{1}{c}{NMI} &
  \multicolumn{1}{c}{ACC} &
  \multicolumn{1}{c}{ARI} &
  \multicolumn{1}{c}{NMI} &
  \multicolumn{1}{c}{ACC} &
  \multicolumn{1}{c}{ARI} &
  \multicolumn{1}{c}{NMI} &
  \multicolumn{1}{c}{ACC} &
  \multicolumn{1}{c}{ARI} &
  \multicolumn{1}{c}{NMI} &
  \multicolumn{1}{c}{ACC} &
  \multicolumn{1}{c}{ARI} &
  \multicolumn{1}{c}{NMI} &
  \multicolumn{1}{c}{ACC} &
  \multicolumn{1}{c}{ARI} \\ \midrule

k-means    & 0.087 & 0.229 & 0.049 & 0.084 & 0.130 & 0.028 & 0.125 & 0.192 & 0.061 & 0.119 & 0.241 & 0.057 & 0.055 & 0.105 & 0.020  \\
SC         & 0.103 & 0.247 & 0.085 & 0.090 & 0.136 & 0.022 & 0.098 & 0.159 & 0.048 & 0.151 & 0.274 & 0.076 & 0.038 & 0.111 & 0.013  \\
AC         & 0.105 & 0.228 & 0.065 & 0.098 & 0.138 & 0.034 & 0.239 & 0.332 & 0.140 & 0.138 & 0.242 & 0.067 & 0.037 & 0.139 & 0.021  \\
NMF        & 0.081 & 0.190 & 0.034 & 0.079 & 0.118 & 0.026 & 0.096 & 0.180 & 0.046 & 0.132 & 0.230 & 0.065 & 0.044 & 0.118 & 0.016  \\
AE         & 0.239 & 0.314 & 0.169 & 0.100 & 0.165 & 0.048 & 0.250 & 0.303 & 0.161 & 0.210 & 0.317 & 0.152 & 0.104 & 0.185 & 0.073  \\
DAE        & 0.251 & 0.297 & 0.163 & 0.111 & 0.151 & 0.046 & 0.224 & 0.302 & 0.152 & 0.206 & 0.304 & 0.138 & 0.104 & 0.190 & 0.078  \\
DCGAN      & 0.265 & 0.315 & 0.176 & 0.120 & 0.151 & 0.045 & 0.210 & 0.298 & 0.139 & 0.225 & 0.346 & 0.157 & 0.121 & 0.174 & 0.078  \\
DeCNN      & 0.240 & 0.282 & 0.174 & 0.092 & 0.133 & 0.038 & 0.227 & 0.299 & 0.162 & 0.186 & 0.313 & 0.142 & 0.098 & 0.175 & 0.073  \\
VAE        & 0.245 & 0.291 & 0.167 & 0.108 & 0.152 & 0.040 & 0.200 & 0.282 & 0.146 & 0.193 & 0.334 & 0.168 & 0.107 & 0.179 & 0.079  \\
JULE       & 0.192 & 0.272 & 0.138 & 0.103 & 0.137 & 0.033 & 0.182 & 0.277 & 0.164 & 0.175 & 0.300 & 0.138 & 0.054 & 0.138 & 0.028  \\
DEC        & 0.257 & 0.301 & 0.161 & 0.136 & 0.185 & 0.050 & 0.276 & 0.359 & 0.186 & 0.282 & 0.381 & 0.203 & 0.122 & 0.195 & 0.079  \\
DAC        & 0.396 & 0.522 & 0.306 & 0.185 & 0.238 & 0.088 & 0.366 & 0.470 & 0.257 & 0.394 & 0.527 & 0.302 & 0.219 & 0.275 & 0.111  \\
ADC       & --     & 0.325 & --     & --     & 0.160 & --  &  &  &     & --     & --     & --     & --     & --     & --    \\
DDC        & 0.424 & 0.524 & 0.329 & --     & --     & --   & 0.371 & 0.489 & 0.267    & 0.433 & 0.577 & 0.345 & --     & --     & --     \\
DCCM       & 0.496 & 0.623 & 0.408 & 0.285 & 0.327 & 0.173 & 0.376 & 0.482 & 0.262 & 0.608 & 0.710 & 0.555 & 0.321 & 0.383 & 0.182  \\
IIC        & --     & 0.617 & --     & --     & 0.257 & --  & -- & 0.610 &  --      & --     & --     & --     & --     & --     & --         \\
PICA & 0.591 & 0.696 & 0.512 & 0.310 & 0.337 & 0.171 & 0.611 & 0.713 & 0.531 & 0.802 & 0.870 & 0.761 & 0.352 & 0.352 & 0.201  \\
CC & 0.705 & 0.790 & 0.637 & 0.431 & 0.429 & 0.266 & 0.746 & 0.850 & 0.726 & 0.859 & 0.893 & 0.822 & 0.445 & 0.429 & 0.274  \\
SPICE & 0.734 & 0.838 & 0.705 & 0.448 & 0.468 & 0.294 & \textbf{0.817} & \textbf{0.908} & \textbf{0.812} & (0.927) & (0.969) & (0.933) & (0.498) & (0.546) & (0.362)  \\
SCAN & 0.796 & 0.861 & 0.750 & 0.485 & 0.483 & 0.314 & 0.703 & 0.818 & 0.661 & -- & -- & -- & -- & -- & --  \\
PCL & \underline{0.802} & \underline{0.874} & \underline{0.766} & \underline{0.528} & \underline{0.526} & \textbf{0.363} & 0.410 & 0.718 & 0.670 & 0.841 & \textbf{0.907} & 0.822 & 0.440 & 0.412 & 0.299  \\ \hline
\textbf{TCL} & \textbf{0.819} & \textbf{0.887} & \textbf{0.780} & \textbf{0.529} &   \textbf{0.531} & \underline{0.357} & \underline{0.799} & \underline{0.868} & \underline{0.757} & \textbf{0.875} & \underline{0.895} & \textbf{0.837} &  \underline{0.623} & \textbf{0.644} & \textbf{0.516}  \\
TCL$_\text{ICH}$ & 0.792 & 0.867 & 0.737 & 0.522 & 0.517 & 0.337 & 0.732 & 0.792 & 0.564 & \underline{0.869} & 0.891 & \underline{0.823} & \textbf{0.624} & \underline{0.639} & \underline{0.503}  \\ \bottomrule
\end{tabular}
}
\label{tab:result}
\end{table*}

For text data, we randomly substitute 20\% words of each text with their top-$n$ suitable words found by the pre-trained Roberta from the Contextual Augmenter Library~\citep{ma2019nlpaug} as weak augmentation, following the setting in SCCL~\citep{SCCL}. The four operations proposed by EDA~\citep{EDA} with a probability of $0.2$ each are adopted as strong augmentations, including SynonymReplacement, RandomInsertion, RandomSwap, and RandomDeletion.

For the two contrastive heads, the dimensionality of ICH is set as $128$ to keep more discriminative information (see ablation study in Section~\ref{sec:decouple}), and the dimensionality of CCH is naturally set to the target cluster number. The temperature parameters are empirically set as $\tau_I=0.5, \tau_C=1.0$ for all datasets. The Adam optimizer with an initial learning rate of $1e{-4}$ and a weight decay of $1e{-4}$ is adopted to jointly optimize the two contrastive heads and the backbone network on image datasets. Since the backbone is pre-trained for text data, we set the learning rate of the optimizer as $5e{-6}$ for the backbone and $5e{-4}$ for two contrastive heads. The model is trained for 1000/500 epochs, followed by 200/100 boosting epochs for the image/text dataset with a batch size of 256. Experiments are carried out on Nvidia TITAN RTX 24G and RTX 3090 on the Ubuntu 18.04 platform with CUDA 11.0 and PyTorch 1.8.0~\citep{PyTorch}.

\subsection{Compared Methods}

The proposed TCL is evaluated on five image datasets and two text datasets. For image clustering, we take 21 representative state-of-the-art approaches for comparisons, including k-means~\citep{Kmeans}, SC~\citep{SC}, AC~\citep{AC}, NMF~\citep{NMF}, AE~\citep{AE}, DAE~\citep{DAE}, DCGAN~\citep{DCGAN}, DeCNN~\citep{DeCNN}, VAE~\citep{VAE}, JULE~\citep{JULE}, DEC~\citep{DEC}, DAC~\citep{DAC}, ADC~\citep{ADC}, DDC~\citep{DDC}, DCCM~\citep{DCCM}, IIC~\citep{IIC}, PICA~\citep{PICA}, CC~\citep{CC}, SPICE~\citep{SPICE}, SCAN~\citep{SCAN}, and PCL~\citep{li2020prototypical}. For those representation-based methods, namely SC, NMF, AE, DAE, DCGAN, DeCNN, and VAE, clustering is achieved by applying the vanilla k-means on the learned features. To ensure the backbone is the same across all recent deep clustering methods, we reproduce SCAN with ResNet34 using its official released code. We would like to point out that SPICE further boosts the clustering performance under a semi-supervised framework, and it uses a deeper and wider ResNet backbone (\textit{e.g.}, WRN37-2) which enjoys a much better feature extraction ability~\citep{SimCLRv2}. Thus, for a fair comparison, here we compare it with its self-trained results which are achieved on ResNet34. Besides, SPICE uses the model pre-trained on ImageNet for ImageNet-10/Dogs (denoted by ``()'' in Table~\ref{tab:result}), while all other methods including ours train the model from scratch.

For text clustering, we compare the proposed TCL with 11 benchmarks, including TF/TF-IDF~\citep{TFIDF}, BagOfWords (BOW)~\citep{BOW}, SkipVec~\citep{SkipVec}, Para2Vec~\citep{Para2Vec}, GSDPMM~\citep{GSDPMM}, RecNN~\citep{ReCNN}, STCC~\citep{STCC}, HAC-SD~\citep{ECIC}, ECIC~\citep{ECIC}, and SCCL~\citep{SCCL}. Similarly, the vanilla k-means is conducted on the extracted features to cluster data for those representation-based methods, including BOW, TF/TF-IDF, SkipVec, Para2Vec, and RecNN.

\begin{figure*}[t]\centering
  \subfigure[0 epoch (NMI=0.152)]{
    \includegraphics[width=0.23\textwidth]{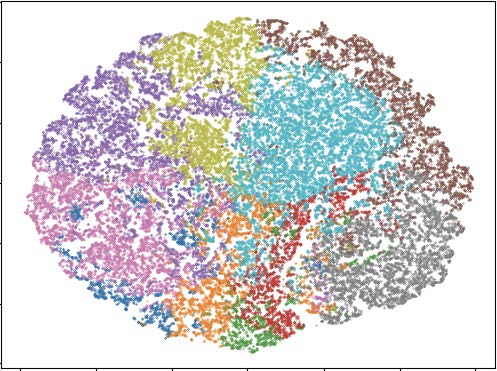}}
  \subfigure[200 epoch (NMI=0.699)]{
    \includegraphics[width=0.23\textwidth]{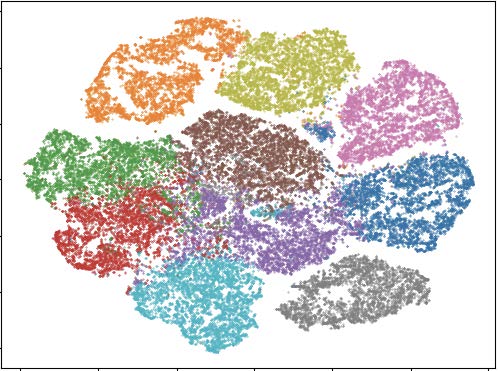}}
  \subfigure[1000 epoch (NMI=0.790)]{
    \includegraphics[width=0.23\textwidth]{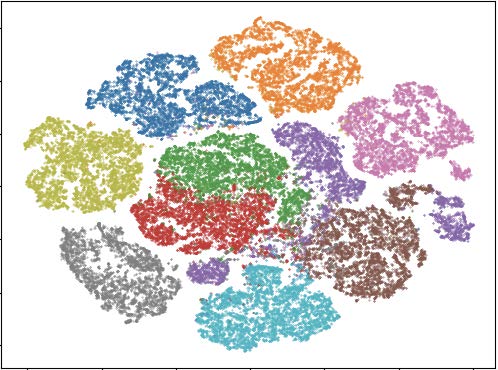}}
  \subfigure[1200 epoch (NMI=0.819)]{
    \includegraphics[width=0.23\textwidth]{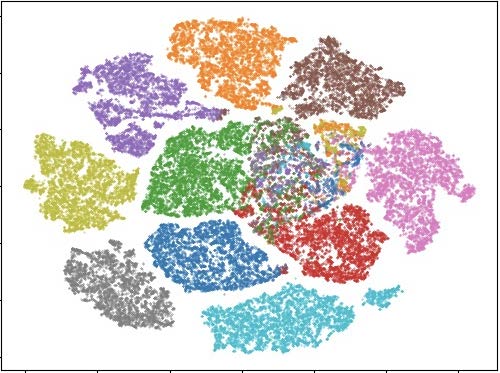}}
  \caption{The evolution of instance features and cluster assignments across the training and boosting stage on CIFAR-10. We perform t-SNE on the features learned by ICH and use different colors to indicate the cluster assignment predicted by CCH.}\label{fig:tsne}
\end{figure*}

\subsection{Evaluation Metrics}

Three widely-used clustering metrics including Normalized Mutual Information (NMI)~\citep{NMI}, Clustering Accuracy (ACC)~\citep{ACC}, and Adjusted Rand Index (ARI)~\citep{ARI} are utilized to evaluate our method. Higher scores indicate better clustering performance.

\subsection{Results}

Both quantitative and qualitative studies are carried out to evaluate the proposed method. Specifically, we compare TCL with state-of-the-art baselines on image and text benchmarks, and visualize the clustering results across the training process.

\begin{table}[t]
\centering
\caption{The clustering performance on two text datasets. The first and second best results are shown in \textbf{bold} and \underline{underline}, respectively. ``TCL$_\text{ICH}$'' refers to clustering results achieved by conducting $k$-means on the ICH features.}

\begin{tabular}{@{}lcccc@{}}
\toprule
Dataset &
  \multicolumn{2}{c}{StackOverflow} &
  \multicolumn{2}{c}{Biomedical} \\ \midrule
Metrics &
  \multicolumn{1}{c}{NMI} &
  \multicolumn{1}{c}{ACC} &
  \multicolumn{1}{c}{NMI} &
  \multicolumn{1}{c}{ACC} \\ \midrule

TF & 0.078 & 0.135 & 0.093 & 0.152 \\
BOW & 0.140 & 0.185 & 0.092 & 0.143 \\
SkipVec & 0.027 & 0.009 & 0.107 & 0.163 \\
TF-IDF & 0.156 & 0.203 & 0.254 & 0.280  \\
Para2Vec & 0.279 & 0.326 & 0.348 & 0.413 \\
GSDPMM & 0.306 & 0.294 & 0.320 & 0.281 \\
RecNN & 0.402 & 0.405 & 0.338 & 0.367 \\
STCC & 0.490 & 0.511 & 0.381 & 0.436 \\
HAC-SD & 0.595 & 0.648 & 0.335 & 0.401 \\
ECIC & 0.734 & 0.787 & 0.413 & \underline{0.478} \\
SCCL & 0.745 & 0.755 & 0.415 & 0.462 \\
\hline
\textbf{TCL} & \underline{0.786} & \textbf{0.882} & \textbf{0.429} & \textbf{0.498} \\
TCL$_\text{ICH}$ & \textbf{0.788} & \underline{0.807} & \underline{0.423}  & 0.470 \\ \bottomrule
\end{tabular}

\label{tab:result_text}
\end{table}

\subsubsection{Comparisons with state of the arts}

The clustering results on five image benchmarks shown in Table~\ref{tab:result} demonstrate the promising performance of TCL. It is worth noting that our method even outperforms SPICE~\citep{SPICE} on the ImageNet-Dogs dataset without ImageNet pre-training, which proves the effectiveness of TCL.

Table~\ref{tab:result_text} shows the clustering results on two commonly-used text datasets. Because existing works seldom use the ARI to evaluate text clustering, here we just adopt NMI and ACC for comparisons. The results show that TCL achieves promising performance on both datasets. We would like to point out that SCCL~\citep{SCCL} is also a contrastive learning based method, which achieves clustering by conducting DEC~\citep{DEC} on the representation learned by the instance-level contrastive learning. The dominance of TCL over SCCL~\citep{SCCL} proves the effectiveness of the proposed twin contrastive learning and confidence-based boosting.

\subsubsection{Evolution of instance features and cluster assignments}

The instance- and cluster-level contrastive learning ought to help the model to learn discriminative representations and predict accurate cluster assignments, respectively. To experimentally study the convergence of TCL, we perform t-SNE~\citep{tSNE} on representations learned by ICH at four timestamps throughout the training and boosting stage, where the cluster assignments predicted by CCH are denoted in different colors. As shown in Fig.~\ref{fig:tsne}, features are all mixed and most instances are assigned to a few clusters at the beginning. As the training progresses, features scatter more distinctly and cluster assignments become more balanced. Finally, more compact and well-separated clusters are achieved with the help of the confidence-based boosting.

\subsection{Ablation Study}

Five ablation studies are carried out to further show the importance of each component in the proposed method. To be specific, the effectiveness of the boosting strategy, the decoupling strategy, and the two contrastive heads are studied. In addition, we investigate the influence of the hyper-parameters in the boosting stage. Besides, different combinations of weak and strong augmentation are tried to verify the effectiveness of the proposed mixed augmentation.

\subsubsection{Effectiveness of the boosting strategy}

To verify the effectiveness of fine-tuning at both the instance and cluster level, we conduct ablation studies by removing one and both of the boosting losses and report the results in Table~\ref{tab:boosting}. The self-labeling loss on CCH is essential in performance boosting because it directly affects the cluster assignments and thus cannot be removed. The results show that both losses improve the clustering performance.

\begin{table}[h!]
\centering
\caption{Effectiveness of the boosting strategy. ``SL'' refers to the self-labeling loss, and ``SCL'' refers to the self-supervised contrastive loss.}
\begin{tabular}{@{}ccccc@{}}
\toprule
Dataset                      & Boost & NMI & ACC & ARI \\ \midrule
\multirow{3}{*}{CIFAR-10}    & None         & 0.790 & 0.865 & 0.752     \\
                             & SL        & 0.805 & 0.878 & 0.770      \\
                             & SL+SCL        &   \textbf{0.819}     &     \textbf{0.887}      & \textbf{0.780}     \\ \midrule
\multirow{3}{*}{ImageNet-Dogs} & None        & 0.518 & 0.549 & 0.381     \\
                             & SL        & 0.562 & 0.600 & 0.441    \\
                             & SL+SCL        & \textbf{0.623}    &   \textbf{0.644}  &  \textbf{0.516}   \\
                             \midrule
\multirow{3}{*}{StackOverflow} & None        & 0.751 & 0.860 & 0.731     \\
                             & SL        &  0.780   &  0.877   & 0.761    \\
                             & SL+SCL        & \textbf{0.786}    & \textbf{0.882}    &  \textbf{0.771}   \\ \bottomrule
\end{tabular}
\label{tab:boosting}
\end{table}

\begin{figure}[t]\centering
    \includegraphics[width=0.9\columnwidth]{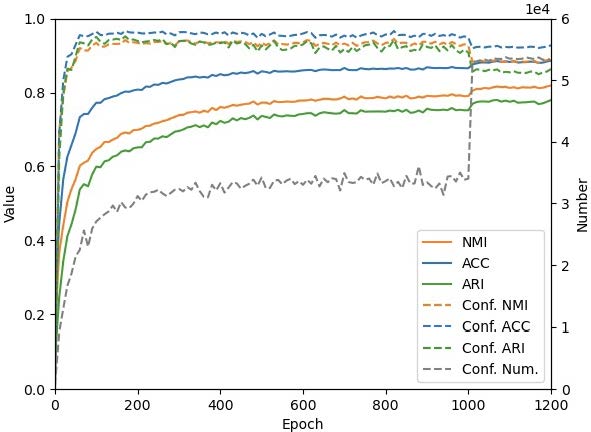}
    \caption{Evolution of clustering performance and confident predictions w.r.t. training epochs on CIFAR-10, where confident predictions are those with $\mathrm{CONF}\geq0.99$.}\label{fig:performance}
\end{figure}

The success of fine-tuning relies on the quality of pseudo labels. As shown in Fig.~\ref{fig:performance}, the confident predictions are more likely to be correct after a period of training, and the model makes more confident predictions as the training progresses. The number of confident predictions significantly increases at the boosting stage due to the self-labeling loss. However, it does not mean that all the confident predictions are selected as pseudo labels since the selection is based on rating instead of an absolute threshold.

Note that as the following ablation studies only influence the training stage, we report their clustering performances without boosting for simplicity.

\subsubsection{Hyper-parameters for pseudo label selection}
\label{sec:hyper-params}

\begin{figure*}[!t]\centering
  \subfigure[threshold $\alpha$]{
    \includegraphics[width=0.3\textwidth]{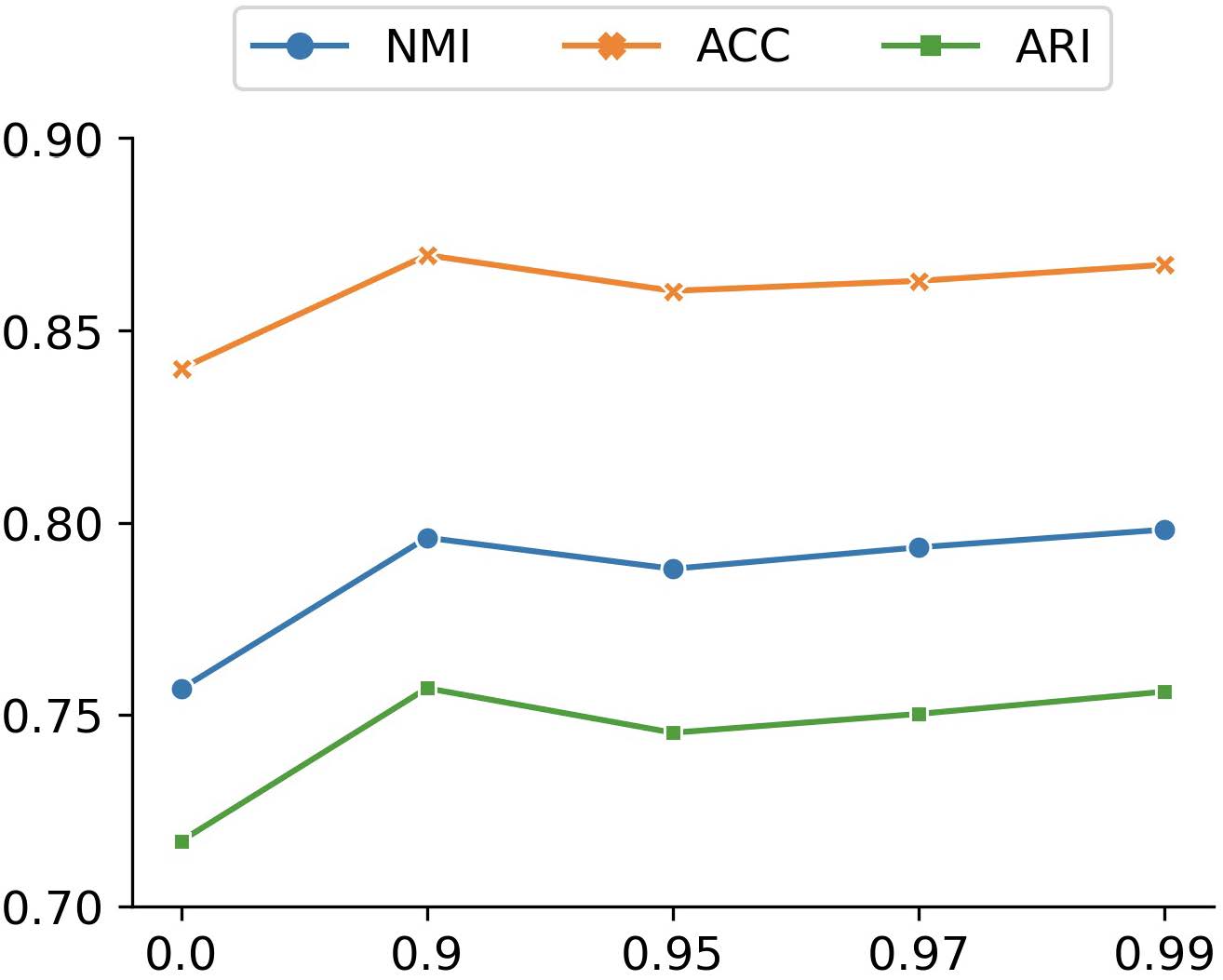}\label{fig:pseudo-a}}
  \subfigure[ratio $\gamma$]{
    \includegraphics[width=0.3\textwidth]{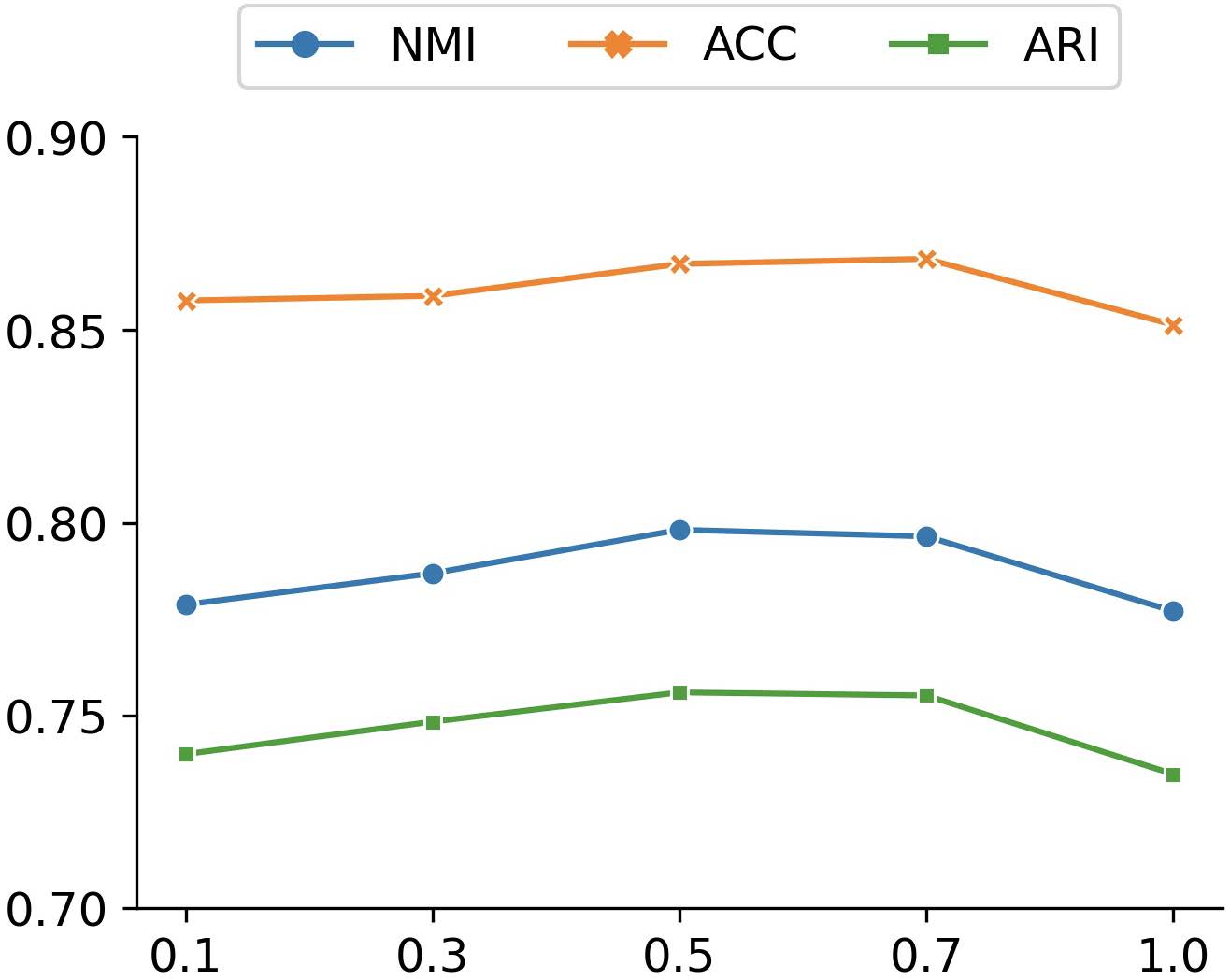}\label{fig:pseudo-g}}
  \subfigure[batch size $N$]{
    \includegraphics[width=0.3\textwidth]{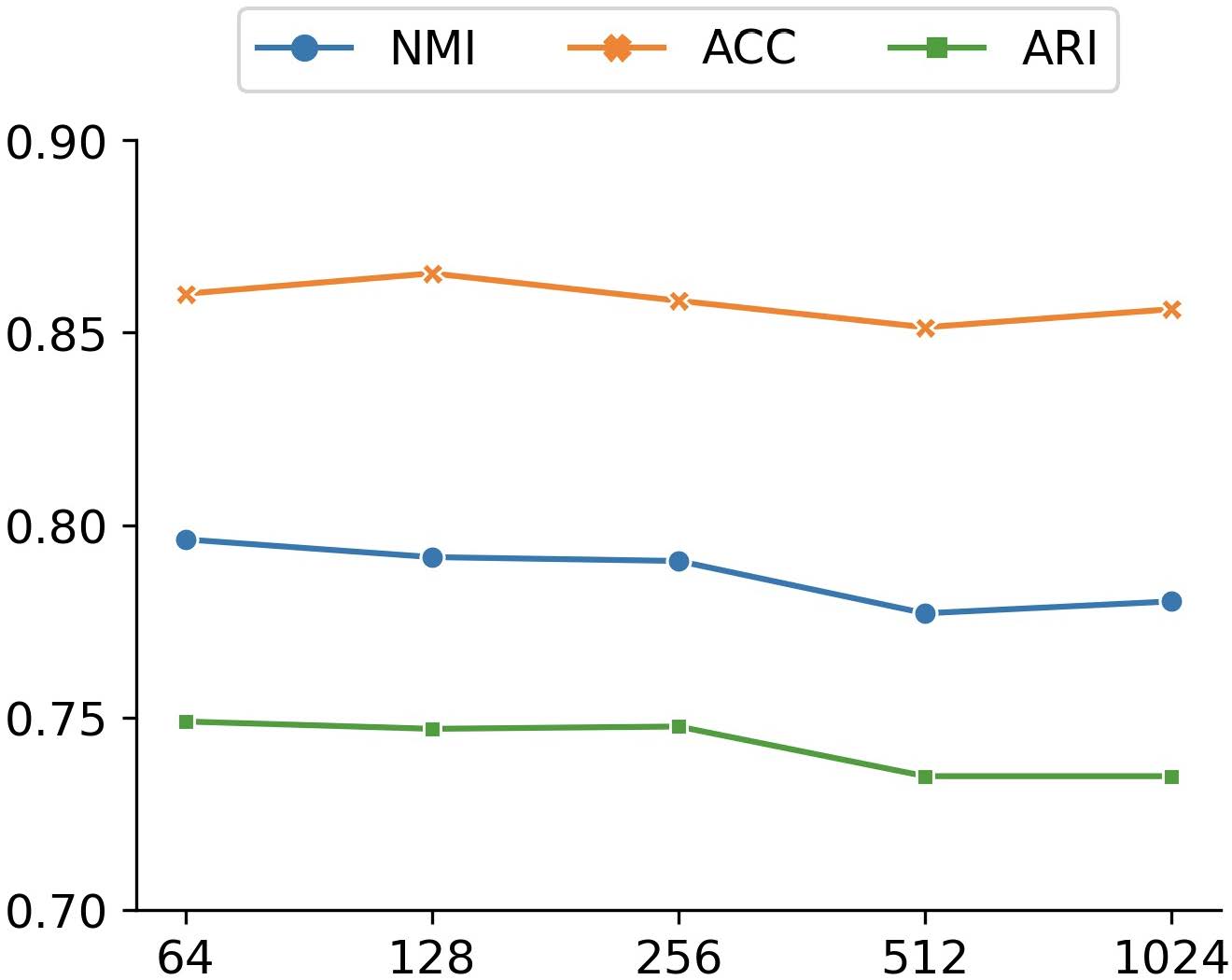}\label{fig:pseudo-n}}
  \caption{Ablation study of different choices of hyper-parameters during the boosting stage on STL-10. One of the hyper-parameters is ablated from the default setting $\alpha=0.99, \gamma=0.5, N=256$ each time.}\label{fig:pseudo}
\end{figure*}

\begin{table*}[!t]
\centering
\caption{Clustering performance gain on CIFAR-10 by introducing the mixed augmentation strategy. $\mathcal{T}$ denotes the weak augmentation, and $\mathcal{T}^\prime$ denotes the strong augmentation.}
\label{tab:gain}
\begin{tabular}{@{}ccccc@{}}
\toprule
Method                          & Augmentation & NMI   & ACC   & ARI     \\ \midrule
\multirow{2}{*}{SimCLR+k-means}  &   $\mathcal{T} + \mathcal{T}$           & 0.699 & 0.782 & 0.616      \\
                             &   $~\mathcal{T} + \mathcal{T}^\prime$           & 0.734 (+0.035) & 0.821 (+0.039) & 0.675 (+0.059)  \\ \midrule
\multirow{2}{*}{TCL}            &  $\mathcal{T} + \mathcal{T}$            & 0.705 & 0.790 & 0.637    \\
                                &  $~\mathcal{T} + \mathcal{T}^\prime$            & 0.790 (+\textbf{0.085}) & 0.865 (+\textbf{0.075}) & 0.752 (+\textbf{0.115}) \\ \bottomrule
\end{tabular}
\end{table*}

Recall that in the boosting stage, we select most confident cluster assignments as pseudo labels based on the threshold $\alpha$ and the ratio $\gamma$. To investigate the influence of two hyper-parameters, we evaluate different choices of them in Fig.~\ref{fig:pseudo}. From Fig.~\ref{fig:pseudo-a} and \ref{fig:pseudo-g}, one could see that the boosting performance is stable across a reasonable range of $\alpha$ and $\gamma$, but degrades when one of the criteria is abandoned (\textit{i.e.}, $\alpha=0.0$ or $\gamma=1.0$). Such a result indicates the robustness and necessity of two pseudo label selection criteria. Besides, since pseudo labels are selected in a batch-wise manner, we also investigate the influence of batch size. To avoid the influence of batch size on contrastive learning, pseudo labels are selected at the beginning of each epoch within mini-batch of different sizes, and the batch size for optimization remains the default (\textit{i.e.}, 256). Fig.~\ref{fig:pseudo-n} shows that the boosting performance is stable in general, but it slightly decreases for large batch sizes. A possible explanation is that for fixed $\alpha$ and $\gamma$, a larger $N$ intrinsically tightens the pseudo label selection. In practice, a smaller $\gamma$ is preferred when the batch size is small and vice versa.

\subsubsection{Different combinations of data augmentation}

As discussed above, we provide a new effective augmentation strategy by mixing weak and strong transformations. To further show its superiority, we validate our model with different combinations of the weak and strong augmentation. The results shown in Table~\ref{tab:augmentation} suggest that for both image and text data, such a mixed augmentation strategy results in the best clustering performance. 

\begin{table}[h!]
\centering
\caption{Importance of data augmentation. $\mathcal{T}$ denotes the weak augmentation, and $\mathcal{T}^\prime$ denotes the strong augmentation.}
\begin{tabular}{@{}ccccc@{}}
\toprule
Dataset                   & Augmentation & NMI   & ACC   & ARI   \\ \midrule
\multirow{3}{*}{CIFAR-10} & $\mathcal{T} + \mathcal{T}$  & 0.712 & 0.805 & 0.648 \\
                          & $~\mathcal{T} + \mathcal{T}^\prime$     & \textbf{0.790} & \textbf{0.865} & \textbf{0.752} \\
                          & $\mathcal{T}^\prime + \mathcal{T}^\prime$          & 0.678 & 0.738 & 0.614  \\ \midrule
\multirow{3}{*}{ImageNet-Dogs} & $\mathcal{T} + \mathcal{T}$ & 0.441 & 0.413 & 0.262  \\
                          & $~\mathcal{T} + \mathcal{T}^\prime$     & \textbf{0.518} & \textbf{0.549} & \textbf{0.381} \\
                          & $\mathcal{T}^\prime + \mathcal{T}^\prime$          & 0.071 & 0.147 & 0.027 \\\midrule
\multirow{3}{*}{StackOverflow} & $\mathcal{T} + \mathcal{T}$ & 0.546 & 0.580 & 0.420 \\
                          & $~\mathcal{T} + \mathcal{T}^\prime$     & \textbf{0.751} & \textbf{0.860} & \textbf{0.731}\\
                          & $\mathcal{T}^\prime + \mathcal{T}^\prime$          & 0.647 & 0.715 & 0.569 \\ \bottomrule
\end{tabular}
\label{tab:augmentation}
\end{table}

Notably, though such an augmentation strategy is seemingly simple, its effectiveness is closely correlated with the proposed TCL framework. Previous works have shown that directly introducing strong augmentation into the contrastive learning framework could lead to sub-optimal performance~\citep{StrongContrastive}. Different from such a conclusion, we show that the mixed augmentation strategy naturally fits the proposed TCL framework. To support the claim, we compare the performance gain obtained by introducing mixed augmentation to the ``SimCLR+k-means'' paradigm and our TCL framework. Table~\ref{tab:gain} shows the clustering performance on CIFAR-10, which demonstrates that our TCL benefits more from the mixed augmentation strategy.

\subsubsection{Effectiveness of the decoupling strategy.}
\label{sec:decouple}

Although the instance- and cluster-level contrastive learning could be directly conducted in the same subspace, in practice, we find that decoupling them into two separate subspaces leads to better clustering performance. Table~\ref{tab:decouple} shows the ablation results, where ``Yes'' denotes that the twin contrastive learning is conducted in two separate subspaces and ``No'' denotes that both contrastive losses are computed in CCH.

Though decoupling the twin contrastive learning into two subspaces improves the performance, it does not contradict our motivation. No matter the two contrastive losses are computed jointly or separately, they act on the same representation $h$, and jointly optimize the network. In fact, it is consistent with the common contrastive learning framework~\citep{SimCLR, MOCO} by stacking projection heads on the representation to compute contrastive loss. The inferior performance of computing two contrastive losses in the same subspace could be attributed to two facts: \textit{i}) the dimension of rows that equals the cluster number is not high enough to contain much information for instance-level contrastive learning (see ablation studies in Table~\ref{tab:dimensionality}); and \textit{ii}) it would lead to a sub-optimal solution since the instance- and cluster-level contrast will influence each other in the same subspace. In brief, the instance-level contrastive learning aims at discriminating different instances instead of clusters, whose optimal solution is thus inferior to the cluster-level contrastive learning.

\begin{table}[!h]
\centering
\caption{Effectiveness of the decoupling strategy.}
\begin{tabular}{@{}ccccc@{}}
\toprule
Dataset                      & Decoupling & NMI & ACC & ARI \\ \midrule
\multirow{2}{*}{CIFAR-10}    & Yes       &   \textbf{0.770}     &     \textbf{0.854}      & \textbf{0.730}    \\
                             & No        &   0.601     &     0.640      & 0.481   \\ \midrule
\multirow{2}{*}{ImageNet-Dogs} & Yes        &   \textbf{0.518}     &     \textbf{0.535}      & \textbf{0.385}\\
                             & No         &   0.343     &     0.368      & 0.214     \\\midrule
\multirow{2}{*}{StackOverflow} & Yes        &   \textbf{0.747}        &  \textbf{0.857}         &    \textbf{0.725} \\
                             & No         &  0.687   &  0.745   &  0.616   \\ \bottomrule
\end{tabular}
\label{tab:decouple}
\end{table}

\begin{table}[!h]
\centering
\caption{Difference choices of the dimensionality of ICH.}
\begin{tabular}{@{}ccccc@{}}
\toprule
Dataset                      & Dimension & NMI       & ACC       & ARI       \\ \midrule
\multirow{5}{*}{CIFAR-10}    & 10                    &   0.770     &     0.854      & 0.730           \\
                             & 32                    &   0.789     &     0.867      & 0.754   \\
                             & 64                 &   \textbf{0.798}     &     \textbf{0.871}      & \textbf{0.760}         \\
                             & 128                  & 0.790 & 0.866 & 0.752      \\
                             & 256          &   0.789     &     0.867      & 0.753           \\ \midrule
\multirow{5}{*}{ImageNet-Dogs} & 15                    &   0.518     &     0.535      &  0.385  \\
                             & 32                    & 0.535 & 0.557 & 0.400 \\
                             & 64                    &   0.529     &     0.551      & 0.389 \\
                             & 128                   & 0.518 & 0.549 & 0.381      \\
                             & 256              &   \textbf{0.541}     &     \textbf{0.562}      & \textbf{0.411}\\
                             \midrule
\multirow{5}{*}{StackOverflow} & 20                    &   0.747        &  0.857         &    0.725       \\
                             & 32                    & 0.749 & 0.858 & 0.726 \\
                             & 64                    & 0.748          &  0.858         &  0.728        \\
                             & 128                   & \textbf{0.751} & \textbf{0.860} & \textbf{0.731}       \\
                             & 256                   & 0.727          & 0.821          & 0.688          \\ \bottomrule

\end{tabular}
\label{tab:dimensionality}
\end{table}

\subsubsection{Importance of twin contrastive heads}

To investigate the effectiveness of the twin contrastive heads (\textit{i.e.}, ICH and CCH), we conduct ablation studies by removing one of them. Since the cluster assignments can no longer be obtained when CCH is removed, we perform k-means on the features learned by ICH instead.

\begin{table}[h!]
\centering
\caption{Effectiveness of two contrastive heads.}
\resizebox{0.49\textwidth}{!}{%
\begin{tabular}{@{}ccccc@{}}
\toprule
Dataset                      & Contrastive Head & NMI            & ACC            & ARI            \\ \midrule
\multirow{3}{*}{CIFAR-10}    & ICH + CCH     & \textbf{0.790} & \textbf{0.865} & \textbf{0.752}        \\
                             & w/o ICH      & 0.633 & 0.658 & 0.522 \\
                             & w/o CCH      & 0.734 & 0.821 & 0.675 \\ \midrule
\multirow{3}{*}{ImageNet-Dogs} & ICH + CCH     & \textbf{0.518} & \textbf{0.549} & \textbf{0.381} \\
                             & w/o ICH    &   0.376     &     0.366      & 0.221            \\
                             & w/o CCH     & 0.504 & 0.535 & 0.336           \\ 
                             \midrule
\multirow{3}{*}{StackOverflow} & ICH + CCH     & \textbf{0.751} & \textbf{0.860} & \textbf{0.731}  \\
                             & w/o ICH      & 0.735 & 0.842 & 0.706          \\
                             & w/o CCH      &  0.666      &   0.732        &   0.516        \\\bottomrule

\end{tabular}
}
\label{tab:contrastive_head}
\end{table}

The results in Table~\ref{tab:contrastive_head} prove the effectiveness of two contrastive heads and show that superior performance is obtained by jointly conducting instance- and cluster-level contrastive learning. Despite the performance improvement brought by CCH, we would like to emphasize that CCH is essential in achieving online clustering as it could directly and independently make cluster assignments for each instance.

\subsubsection{Over-clustering experiments}
It is highly expected that the clustering methods could be robust to different choices of the target cluster number. Therefore, we conduct over-clustering experiments by doubling the ground-truth cluster number (and even 100 classes for CIFAR-100). In the experiments, all parameters except the CCH dimension remain the same as the default. As the Hungarian matching is not practical to find a many-to-one mapping for over-clustering evaluation, we adopt the majority voting mechanism as an alternative. Namely, we assign all samples of one cluster to the majority ground-truth class among the cluster, being consistent with the criterion used in SCAN~\citep{SCAN}. The results in Table~\ref{tab:over-cluster} demonstrate that TCL is robust against different target cluster numbers.

\begin{table}[!h]
\centering
\caption{The robustness of TCL against different choices of the target cluster number.}
\label{tab:over-cluster}
\resizebox{0.49\textwidth}{!}{%
\begin{tabular}{@{}ccccc@{}}
\toprule
Dataset                    & Target cluster number & NMI   & ACC   & ARI   \\ \midrule
\multirow{2}{*}{CIFAR-10}  & Standard (10)         & \textbf{0.790} & \textbf{0.865} & \textbf{0.752} \\
                           & Over-cluster (20)     & 0.759 & 0.846 & 0.718 \\ \midrule
\multirow{3}{*}{CIFAR-100} & Standard (20)         & 0.477 & 0.481 & 0.303 \\
                           & Over-cluster (40)     & \textbf{0.520} & \textbf{0.579} & \textbf{0.380} \\
              & Over-cluster (100)    & 0.493 & 0.566  & 0.359 \\  \bottomrule
\end{tabular}
}
\end{table}

Essentially, the effect of over-clustering is to increase the intra-class variance. Such behavior is more favorable when the data is intrinsically fine-grained (\textit{i.e.}, composed of several subclasses). Forcing the model to produce more fine-grained partitions could break the cluster structure when the data is intrinsically coarse-grained, which explains the performance drop on CIFAR-10. On CIFAR-100 instead, the over-clustering significantly improves the clustering performance. However, one may note that over-clustering with 100 clusters is slightly worse than that with 40 clusters. Such a performance drop could attribute to the insufficiently large batch size. To correctly represent clusters in CCH, it is necessary to include a reasonable number of samples in each cluster. With a batch size of 256, there are only two samples in each cluster on average. And some clusters might even be empty in some mini-batches, which would harm the cluster-level contrastive learning, leading to inferior clustering performance.

Note that in the proposed TCL framework, the target cluster number needs to be manually set. In practice, when the intrinsic cluster number is unknown, one could adopt some cluster number searching or community detection methods like X-means~\citep{xmeans} and Louvain~\citep{louvain} on the ICH output to estimate the cluster number. Moreover, as our TCL is robust to different choices of cluster numbers, the estimation does not necessarily need to be very precise. And one could adjust the dimensionality of CCH to cluster data under different resolutions depending on the practical needs.

\subsubsection{Scaling-up to ImageNet}
To see how TCL scales to large datasets with much more instances and clusters, we further conduct experiments on the ImageNet dataset. The training split with 1,281,167 images is used for both training and evaluation. As discussed above, cluster-level contrastive learning requires a large batch size to ensure a reasonable number of samples exist in each cluster. Thus, we set the batch size to 4,096 on ImageNet with 1,000 classes. However, we encounter the ``out of GPU memory'' problem even on eight RTX 3090 GPUs with such a large batch size. As a solution, we inherit the ResNet50 model learned by MoCo v2~\cite{MOCOV2} and freeze the first two blocks. In other words, we only optimize the last two blocks of ResNet50 and two contrastive heads. Due to the heavy computational burden, we train and boost the model for 100 and 20 epochs, respectively. We set $\tau_I=\tau_C=0.2$ for training and use the default $\alpha=0.99$ and $\gamma=0.5$ for boosting. In the evaluation, we choose MoCo v2 (the model we inherited) as a baseline by conducting k-means on the extracted features. The clustering performance is shown in Table~\ref{tab:ImageNet}, which proves the effectiveness of the twin contrastive learning framework, the boosting strategy, and the scalability of our method.

\begin{table}[h]
\centering
\caption{TCL scales up to ImageNet-1K with much more images and clusters. $^\dag$means without the boosting stage.}
\label{tab:ImageNet}
\begin{tabular}{@{}lccc@{}}
\toprule
Metrics               & NMI                  & ACC                  & ARI                  \\ \midrule
MoCo (inherited model)              &   0.6186                   &   0.3047                   &    0.1428                  \\
TCL$^\dag$       &        0.6332              &     0.3160                 &     0.1901                 \\
TCL                   & \textbf{0.6711} & \textbf{0.3789} & \textbf{0.2656} \\\bottomrule
\end{tabular}
\end{table}

\begin{figure}[h]\centering
  \subfigure[ICH temperature $\tau_I$]{
    \includegraphics[width=0.23\textwidth]{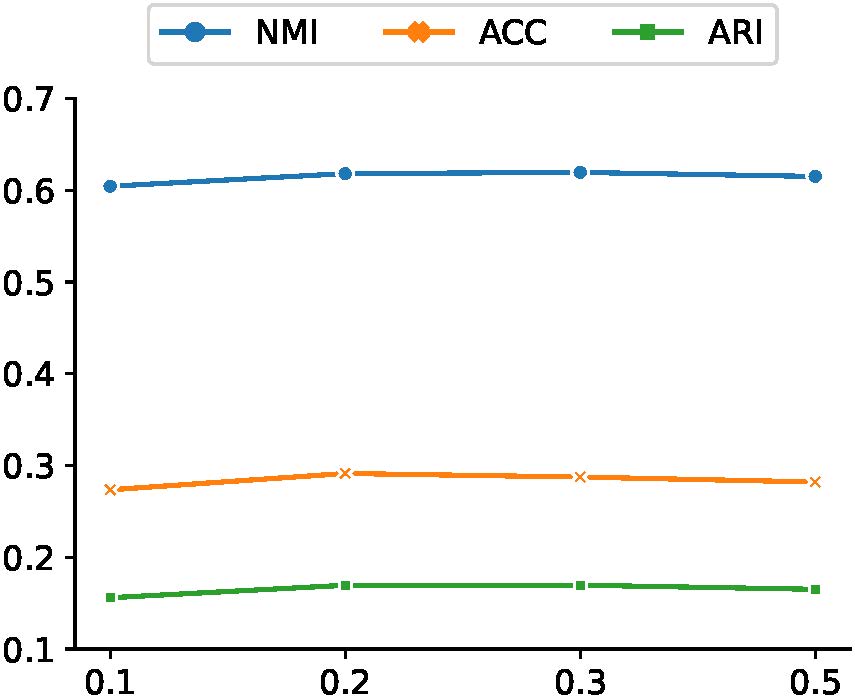}\label{fig:temperature-I}}
  \subfigure[CCH temperature $\tau_C$]{
    \includegraphics[width=0.23\textwidth]{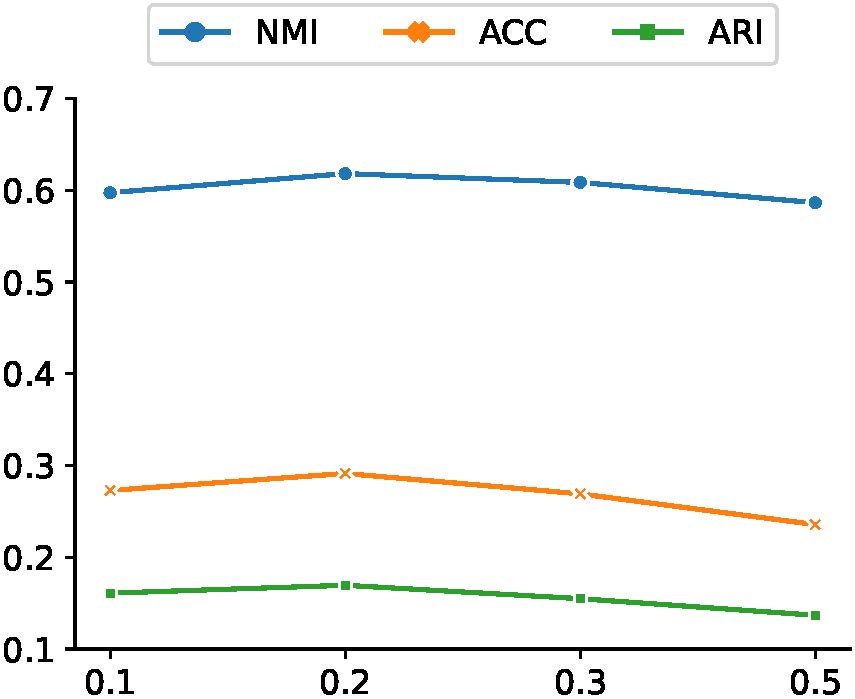}\label{fig:temperature-C}}
  \caption{Ablation study of different choices of temperature $\tau_I$ and $\tau_C$ in TCL, trained for 50 epochs on ImageNet. One of the hyper-parameters is ablated from $\tau_I = \tau_C = 0.2$.}\label{fig:temperature}
\end{figure}

\begin{figure}[h]\centering
  \subfigure[threshold $\alpha$]{
    \includegraphics[width=0.23\textwidth]{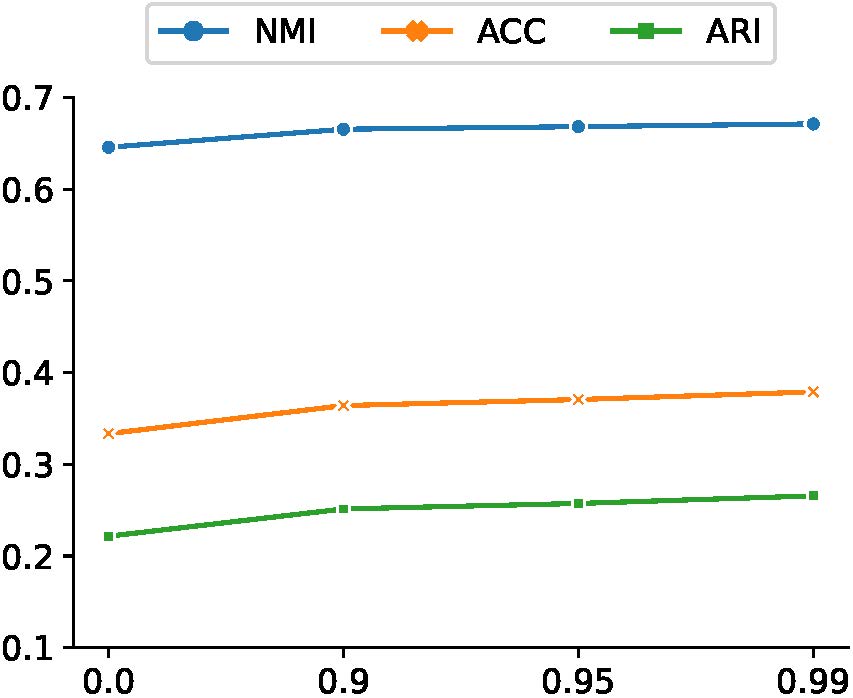}}
  \subfigure[ratio $\gamma$]{
    \includegraphics[width=0.23\textwidth]{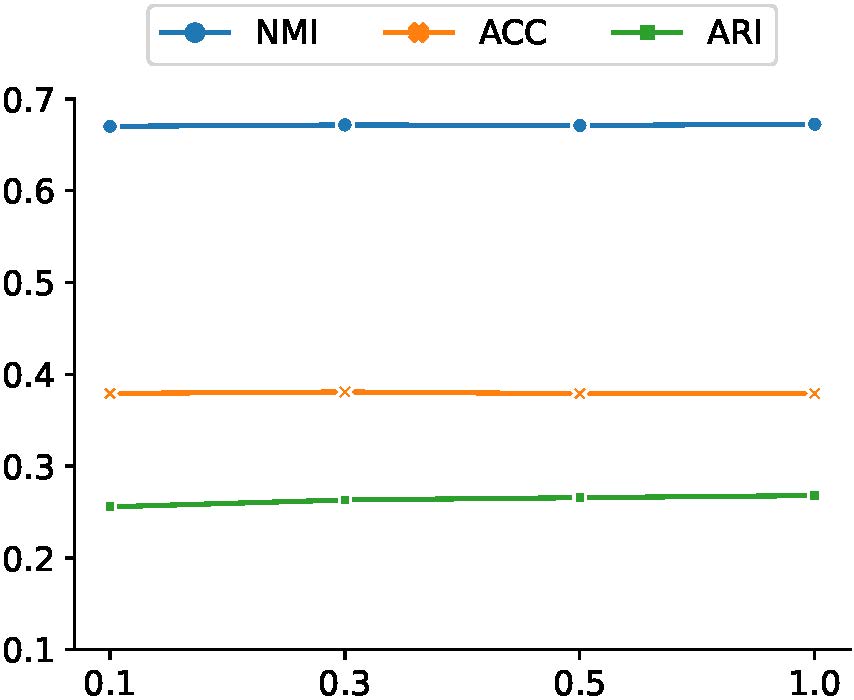}}
  \caption{Ablation study of different choices of $\alpha$ and $\gamma$ during the boosting stage on ImageNet. One of the hyper-parameters is ablated from the default setting $\alpha = 0.99$ and $\gamma = 0.5$.}\label{fig:alpha_gamma}
\end{figure}

As suggested in SimCLR~\citep{SimCLR}, a smaller temperature $\tau_I$ in instance-level contrastive loss could speed up convergence with a large batch size. To investigate the influence of $\tau_I$ and $\tau_C$ in our TCL framework when scaling up to large batch size and cluster number, we ablate these two parameters in Fig.~\ref{fig:temperature}. As shown, a proper choice of $\tau_I$ could slightly improve the performance. More importantly, a smaller temperature $\tau_C$ is also preferred in cluster-level contrastive learning with a large cluster number, since it could sharpen the cluster assignments to obtain a more discriminative cluster representation. We further conduct ablation studies on two boosting hyper-parameters $\alpha$ and $\gamma$. The results in Fig.~\ref{fig:alpha_gamma} show that the threshold $\alpha$ plays a more important role on ImageNet compared with the ratio $\gamma$. This is because all samples are likely to meet the ratio criterion eventually, as there are only four samples in each cluster on average.

\section{Conclusion}

Based on the observation that the rows and columns of the feature matrix intrinsically correspond to the instance and cluster representations, we propose an online deep clustering method termed Twin Contrastive Learning (TCL). By dually conducting contrastive learning at the instance and cluster level, TCL simultaneously learns representations and performs clustering. In addition, to alleviate the influence of intrinsic false negative pairs in the instance-level contrastive learning and to rectify cluster assignments, we propose a confidence-based boosting strategy to further improve the performance by selecting some pseudo labels to fine-tune the twin contrastive learning. Extensive experiments demonstrate the effectiveness of TCL on five image benchmarks and two text datasets.

In the future, we plan to take a deeper look at the influence of different augmentations on contrastive learning. Furthermore, it is worthwhile to explore how to better utilize pseudo labels to identify false negative pairs for improving contrastive learning. 

\begin{acknowledgements}
The authors would like to thank the Associate editor and reviewers for the constructive comments and valuable suggestions that remarkably improve this study. 
This work was supported in part by the National Key R\&D Program of China under Grant 2020YFB1406702; in part by NFSC under Grant 62176171, U21B2040, and U19A2078;  and in part by Open Research Projects of Zhejiang Lab under Grant 2021KH0AB02. 
\end{acknowledgements}

%
%

\bibliographystyle{spbasic}      
\bibliographystyle{spmpsci}      
\bibliographystyle{spphys}       
\bibliography{reference.bib}   

\end{document}